# A Deep Learning Approach to Integrate Human-Level Understanding in a Chatbot

Afia Fairoose Abedin[1], Amirul Islam Al Mamun[1],
Rownak Jahan Nowrin[1], Amitabha Chakrabarty[1],
Moin Mostakim[1] and Sudip Kumar Naskar[2]

[1]Department of Computer Science and Engineering,
Brac University, Dhaka,Bangladesh
[2]Department of Computer Science and Engineering,
Jadavpur University, Kolkata, India

## Abstract

*In recent times, a large number of people have been involved in establishing their own businesses. Unlike humans, chatbots can serve multiple customers at a time, are available 24/7 and reply in less than a fraction of a second. Though chatbots perform well in task-oriented activities, in most cases they fail to understand personalized opinions, statements or even queries which later impact the organization for poor service management. Lack of understanding capabilities in bots disinterest humans to continue conversations with them. Usually, chatbots give absurd responses when they are unable to interpret a user's text accurately. Extracting the client reviews from conversations by using chatbots, organizations can reduce the major gap of understanding between the users and the chatbot and improve their quality of products and services.Thus, in our research we incorporated all the key elements that are necessary for a chatbot to analyse andunderstand an input text precisely and accurately. We performed sentiment analysis, emotion detection, intent classification and named-entity recognition using deep learning to develop chatbots with humanistic understanding and intelligence. The efficiency of our approach can be demonstrated accordingly by the detailed analysis.*

## 1. Introduction

In recent times, a large number of people are involved in establishing their own business. But it is tremendously tough to stay updated with technology and hold a place in the market full of competition. Chatbot is one of the most advanced features incorporated in most organizations or online platforms today. A customer care team has a lot of constraints to working hours, responding to multiple at the same time and also efficiency. Conversational agents definitely solve all the problems. Unlike humans, it doesn't get tired or make delays to reply. Chatbots handle similar questions easily, help firms in advertising their brands in a very cost-effective way and also help organizations to overcome their drawbacks by analysing customer feedback [1]. Not only in business, but also applications of chatbots are emerging in medical fields as well. Chatbot applications are developed to imitate psychiatrist techniques to uplift a person's mood or reduce stress. Completely replacing a human might be quite challenging, but conversational agents can reduce a lot of human effort. There are many conventional AI-bots (Artificial Intelligence Bots) in the market which generate fixed responses and thus can be monotonous at times for the user as it gives redundant replies. Usually, it identifies the most similar keywords and responds to the closely matched text from its database. Eventually, the replies become inconsistent and meaningless most of the time. Even if we talk about one of the first stable virtual assistants, SIRI, which is developed by Apple, has a vast range of features helping users to manage calendars, make calls, schedule meetings and what not. However, if you have ever tried to talk with it more interactively, it often fails to understand and most frequently replies *"I am sorry, I couldn't understand. Could you try again?"*. Furthermore, using a bot in the psychological field needs to be even more accurate and human-like by giving the vibe of human-connection and empathy. That's why even today most users prefer human agents to solve their problems. For a virtual agent to respond like a human, it needs to understand the problem of the user and provide sensible replies to humans [2]. So, the aim of our research is to analyse the various features that are required by a chatbot to gain a human-like understanding of text. We are trying to develop a hybrid model that will understand the sentiment, emotion, intent and named-entities of the user's text. Once this problem is solved, it can be used in every sector for personalised conversations and this data can be very helpful to the organizations whether to help a mental health patient or to improve various services and products sold. Even universities can use it for finding out the exact problem that a student is going through by analysing the bot-human conversation. We chose particularly deep learning because DL models help to extract the meaning of mixed contrary complex sentences given by a user more accurately. The layers in DL models creates a network that helps bots to learn accurately on their own which normal ML bots fail to do [3]. Hence, we believe our research will help to improve chatbot models definitely and open the doors of vast and effective bot-applications.

## 2. RELATED WORK

To know about previous works on sentimental analysis, we read the research paper [4], where text mining techniques were applied to find sentiment and emotion from twitter dataset. Words from twitter were compared with the sentiment file using Bayes algorithm and given the corresponding sentiment. However, it is unable to predict sentiment of any word apart from the words stored in the database and therefore, we get a scope here to improve the drawbacks of their work. Also, an interesting Bangla intelligent social robot was introduced in the paper [5] which communicates in Bangla analysing sentiments with the help of machine learning algorithms. The authors of paper [5] have refined a huge amount of data to fit in their machine learning algorithm. Fictional conversation extracted from raw movie script were used as the metadata collection of the corpus. Naive Bayes classification algorithm was used to train the dataset. To work on context understanding of a chatbot, we went through the paper [6], which was a chatbot application based on NLP along with emotion recognition. The authors added emotion embedding vectors in the output layer of RNN. The system uses 4 levels of hierarchy to understand natural language input sentences and recognize the user's emotion. The response generation is done through collecting, analysing and integrating input dialogues consisting of text. Then document intention extraction is done from the text by neural network. The research paper [7] stated how deep learning can playa significant role in developing realistic chatbots by "mimicking human characteristics" and showed comparison between conversations by using different Neural Network models like Uni-directional LSTM, Bi-directional LSTM, Bi-directional LSTM with attention mechanism. The authors implemented a hybrid chatbot using rule-based and generative models together to makea more meaningful response. According to paper [8], to understand the user's current state of mind and wellness in depth, emotion recognition is helpful which is a detailed version of sentimental analysis. The paper further discusses the various datasets to perform emotion recognition and tabulated a list of already implemented emotion recognition and their limitations. The paper [9] detects emotion from a text using Bi-directional neural networks and tensor flow libraries. It describes how sentiment and emotion are interconnected and can be used to give better responses. The dataset is cleaned, processed and tokenized to fit to Bi-RNN model for training. The model had two parameters, one for sentimental analysis and another to generate dialogue. The work wasn't very efficient as they simply labelled the emotion as happy if sentiment is positive and sad if sentiment is negative. It did not classify the emotions and thus, failed to actually signify the improvement that could be achieved by classifying detailed emotion from the polarity of sentiments. The model played a happy song if the sentiment was positive and vice versa. However, our work completed the drawbacks of this paper by distributing emotion into different classes. Moreover, the authors of the paper [10] surveyed the effectiveness of implementing Named-Entity Recognition by deep learning. The author gave the information of named-entity datasets and where they can be used. The paper stated the different evaluation metrics of NER such as type, exact match, relaxed F1 and strict F1 metrics etc to analyse the performance and accuracy of NER. The survey analysed knowledge-based systems, unsupervised systems, feature-engineered systems, feature-inference neural network systems, word level architectures, character-level architectures, character+word level architecture and lastly, character+word+affix model. After evaluation it was noticed that neural network models outperformed all other applied models mentioned above. Furthermore, the paper [11] proposed their research on Dialogue intent classification by hierarchical LSTM which is an open-source library. Through this a model will be able to recognise the reason behind the uttered sentence. The experiment was performed on a Chinese ecommerce site. The authors compared accuracy between basic LSTM, HLSTM and their hybrid model HLSTM + Memory. By going through all the knowledgeable and qualitative research, we gathered the loopholes of all the papers that needed improvement as well we ensured that our working domain, deep learning, is the right choice for incorporating all the works we studied so far. In our literature review, we have seen the development of models by sentimental analysis or emotion or intent or named-entity. However, there hasn't been yet an analysis that clusters all the components together. Thus, we decided to observe the different performances as well as challenges while using all the features together.

## 3. RESEARCH METHODOLOGY

We started our research by analysing a commonly used feature - **sentiment**, which is incorporated in most customer service chatbots to help the company rate their products. It is also essential in various movie reviews or story book feedback. We implemented various deep learning models from simple neural networks to LSTM and GRU. Based on accuracy, we selected LSTM as the best model for our sentimental analysis. Next, through our study, we came up to classify sentiment into more meaningful expression that is **emotion detection**. We used 7 emotions of joy, sadness, disgust, shame, guilt, anger and fear to identify the specific mental state of the user. We applied Bi-LSTM and Bi-GRU for predicting the correct emotion. Next, we decided to bring out the **intent** of the conversation. We need to know the intent of the text in order to understand what the user wants or what the user needs. Similar to emotion detection, we used Bi-LSTM and Bi-GRU to experiment the intent classification. For getting more details about intent, we performed **named-entity recognition** as well. Named-entity recognition is important because it identifies the person, place, organization, region or even the specific date or time the user is talking about. For example- *"What is the time in Bangladesh now?"* Here *"knowing time"* is the intent and *"Bangladesh"* is the entity. Like emotion detection and intent classification, we used Bi-LSTM and Bi-GRU to recognize named-entity. As we have performed different analysis on different human understanding features it was quite impossible to find one single dataset to incorporate all the analysis. Thus, all our feature analysis was executed by independent datasets. As our domain was deep learning, we performed all our analysis within this field. Despite the unavailability of one dataset, we tried to integrate the analysed models so that all the features can be extracted from a single text. To give accurate results, Deep Learning bots need continuous training. So, it was quite obvious to get incorrect results initially. However, through our accuracy of models, estimations and various research we proposed, a text analysis structure that will give a chatbot multimodal understanding. The following Figure 1 shows the workflow of our executed analysis.

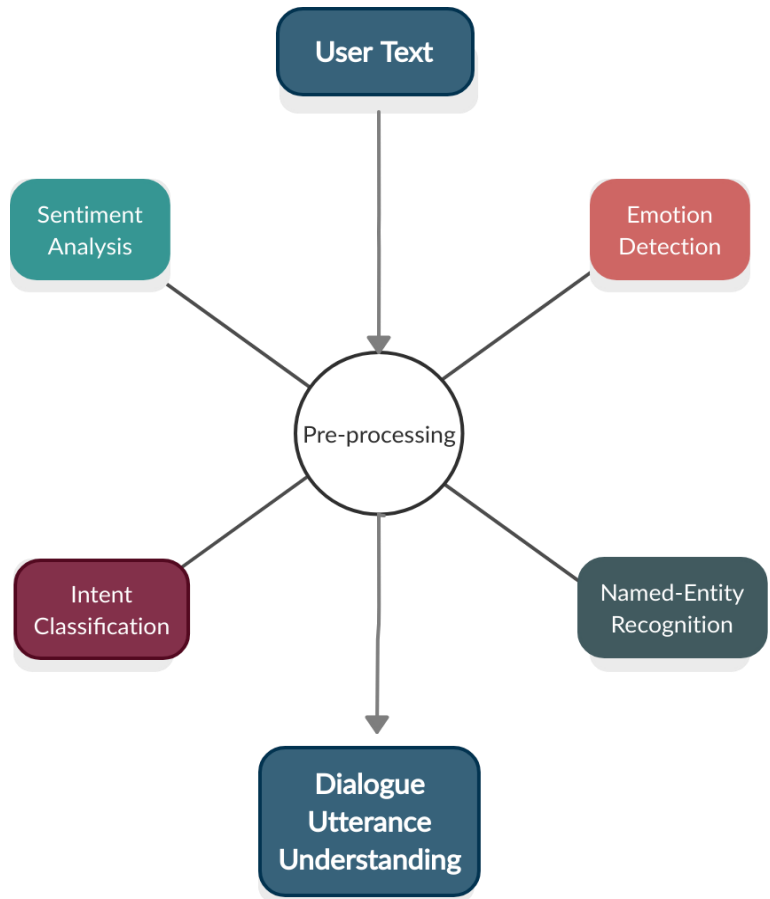

Figure 1.  Analysis work flow for dialogue utterance understanding

## 4. EXPERIMENTAL ANALYSIS AND RESULTS

### 4.1. Sentiment Analysis

For the analysis of information in texts where opinions are highly unstructured and are either positive or negative sentiment analysis is essential [12]. For humans, it is very simple to identify positive and negative sentences. But, a chatbot can never understand the polarity of a text if it is not trained to identify it. As our goal is to create a human-like understanding of text in a chatbot, it is important to understand the sentiment of the content. We applied a simple neural network, GRU and LSTM to perform sentiment analysis on our sentiment datasets. For our work, we used two datasets for analysing our performance in different neural network models with different datasets.

#### 4.1.1. Data Processing

The first dataset is IMDB dataset of movie review which consist of two columns [13]. The second dataset is twitter sentiment analysis dataset which has two columns as well [14]. After cleaning and refining the two datasets, we split them to test and train sets in the ratio 1:3 respectively. Then, we tokenized the datasets separately to fit into texts and further converted texts into sequence. In Figure 2, the whole workflow process of sentiment analysis is described in depth.

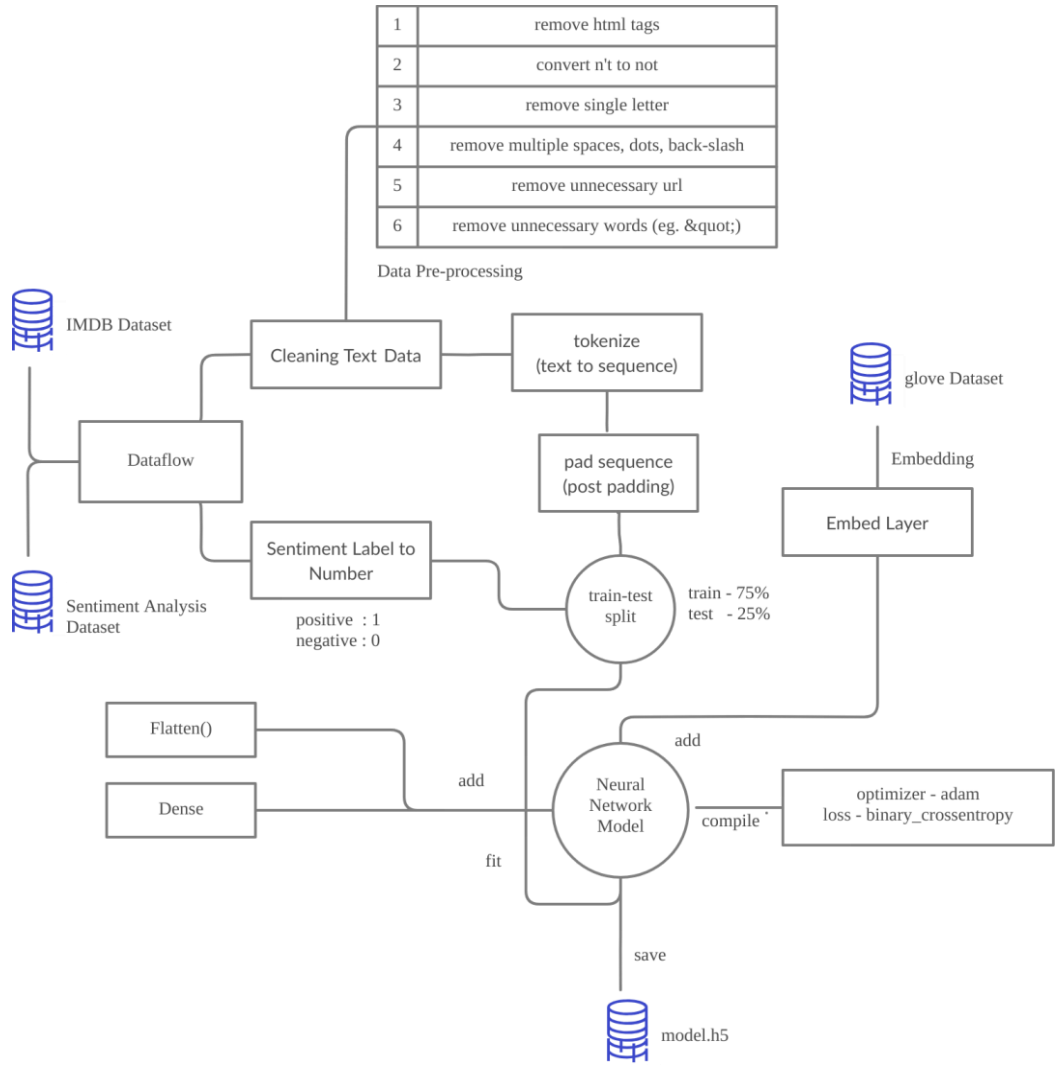

Figure 2.  Training Work Flow for Sentiment Analysis

### 4.1.2. Method

For capturing the context of a word in the datasets, we used Stanford's Glove word embedding which is an unsupervised learning algorithm for obtaining vector representations for words [15]. As different neural network models give different results and accuracy, we tried to implement a number of them such as Simple Neural Network, LSTM and GRU. In our models we have used sigmoid activation function and Adam optimizer. It was important to keep our batch size and epoch moderate to get rid of the overfitting problem. So, we kept the batch size at 128 and epoch to 20 for all our models. We kept the validation split to 0.2. After that we feed the two datasets into the model following the above parameters and demonstrated the different accuracy results of the two datasets.

### 4.1.3.  Result

In Table 1, we showed comparison between the accuracy and loss of different DL models. From our performance analysis we found LSTM gave the highest accuracy in both test and train dataset. The train accuracy in dataset 1 and 2 are 89% and 86% respectively and test accuracy are 80% and 71% respectively. In other models the test accuracy dropped significantly compared to training accuracy.

Table 1. Accuracy for training and validation data of different Datasets and different Deep-Learning Models for Sentiment Analysis

| model | Dataset | Train Data | | | |
|---|---|---|---|---|---|
| | | loss | accuracy | loss | accuracy |
| Simple Neural Network | IMDB Dataset | 0.37 | 0.86 | 0.82 | 0.68 |
| | Twitter Sentiment | 0.54 | 0.74 | 0.66 | 0.65 |
| Gated Recurrent Unit | IMDB Dataset | 0.32 | 0.91 | 0.95 | 0.79 |
| | Twitter Sentiment | 0.74 | 0.80 | 1.19 | 0.68 |
| Long-Short Term Memory | IMDB Dataset | 0.33 | 0.89 | 0.65 | 0.80 |
| | Twitter Sentiment | 0.43 | 0.86 | 0.92 | 0.71 |

### 4.1.4. Performance analysis

Finally in Figure 3, are some sentences given as input to make some predictions whether the sentences were classified correctly as positive or negative. We have set the range of confidence level of a sentence from 0 to 1. The sentences which have confidence levels greater than or equal to 0.5 are given a positive sentiment label and if it is less than 0.5, a negative sentiment label is given. The initial input text was *"Hey, you are a good bot"* and the bot was able to predict the sentiment 'positive' accurately. Usually, it is easier to extract sentiments from simple sentences, so we gave some complex sentences like *"You're a good bot but the behaviour that you did yesterday was not fair"*. Our bot was even successful in catching the right sentiment despite the complexity of text. It was able to consider the sentiment of both the clauses in the sentence and extract the sentiment with higher probability.

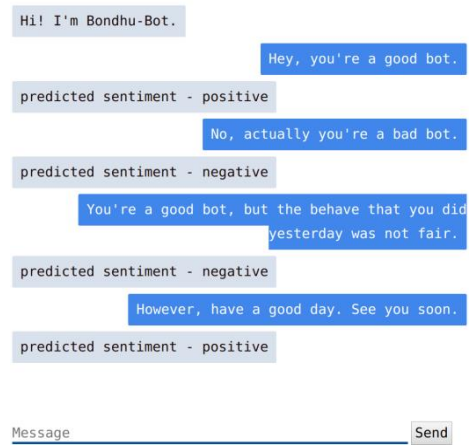

Figure 3.  Application of Sentiment Analysis

## 4.2. Emotion Detection

Sentiment analysis has helped us only to learn the attitude of a person from a user's input. For example- *"What a dream it was! I couldn't sleep at all"*, here by sentimental analysis we can classify this text as a negative sentence, however, this is not enough to understand a person's situation, we need to identify the exact mental state of the user. Let's analyse the above example again. If we closely analyse the text, we can see the person uttered the sentence out of some *'sadness'* and some *'fear'*. This is the exact state of mind we need to capture in our chatbot. Emotions are very subtle and ambiguous in a text which makes it really tough to detect. For that reason, we have focused on using neural network models in this paper to capture the emotion of the user.

### 4.2.1. Data Processing

The dataset used in this model is the International Survey on Emotion Antecedents and Reactions (ISEAR) database created by Klaus R. Scherer and Harald Wallbott which consists of 7665 sentences labelled by seven emotions like joy, sadness, fear, anger, guilt, disgust, and shame [16]. The dataset contains two columns: emotion label and text. As shown in Figure 4, we split all the phrases into smaller parts called tokens using Keras tokenizer. For stemming we used the Lancaster stemmer of NLTK (Natural language toolkit).

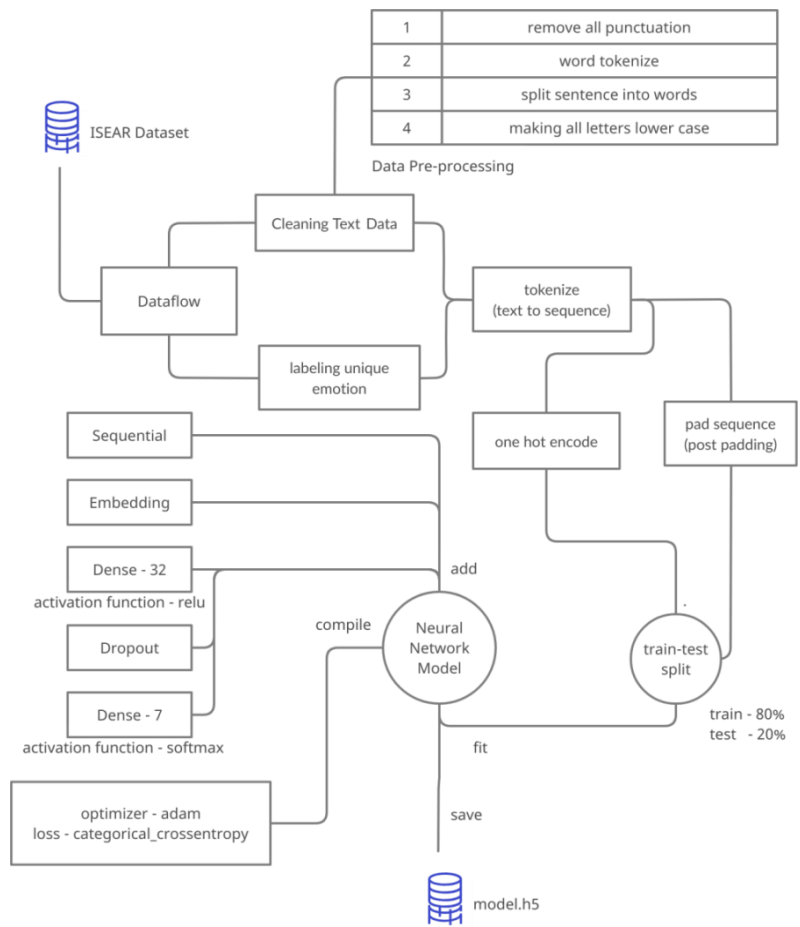

Figure 4.  Training Work Flow for Emotion Detection

**4.2.2. Method**

For the analysis, we have used Bi-GRU and Bi-LSTM and did some comparison which works best for our analysis. We used ReLU (Rectified Linear Unit) and softmax as our activation function. Finally, we have set the optimizer to Adam for accurate results. Moreover, to save the model so that it can be loaded later we used the model checkpoint call-backs of Keras call-back library. Checkpointing the neural network model is essential as checkpoints are the weight of the model and these weights can be used to make better predictions. The livecloss plot tool was applied so that we could get the accuracy-loss plot. We took a batch size of 32 and epoch was kept at 100 for both the models so that the algorithm sees the entire dataset 100 times.

**4.2.3. Result**

Table 3 shows the accuracy and loss results between Bi-LSTM and Bi-GRU. Comparing the results, we can say Bi-GRU gives the highest training accuracy of 93%.

Table 3. Accuracy for training and validation data of Deep-Learning Models for EmotionDetection

| Model | Training | | Validation | |
|---|---|---|---|---|
| | loss | accuracy | loss | accuracy |
| Bi-directional LSTM | 0.350 | 0.875 | 1.602 | 0.684 |
| Bi-directional GRU | 0.173 | 0.936 | 1.556 | 0.437 |

**4.2.4. Performance analysis**

In Table 4, we showed the background values of how an input sentence is classified on the basis of confidence level for each emotion. The emotion with the highest confidence level will be the output. Moreover, few examples with corresponding outcomes are given in Figure 5. Despite the exact words from emotion categories are not directly mentioned in the text yet we can see our model was able to accurately identify the accurate state of mind. For example- it is a matter of "guilt" if I am not being to help my thesis team as it is my responsibility.

Table 4. Predicted confidence level of different emotions

| Text: I did not help my thesis team enough. | |
|---|---|
| **Emotion** | **confidence** |
| guilt | 0.49738222 |
| shame | 0.36354083 |
| anger | 0.05595107 |
| sadness | 0.037771154 |
| fear | 0.022115987 |
| disgust | 0.015596252 |
| joy | 0.0076424023 |
| **predicted result** | **guilt** |

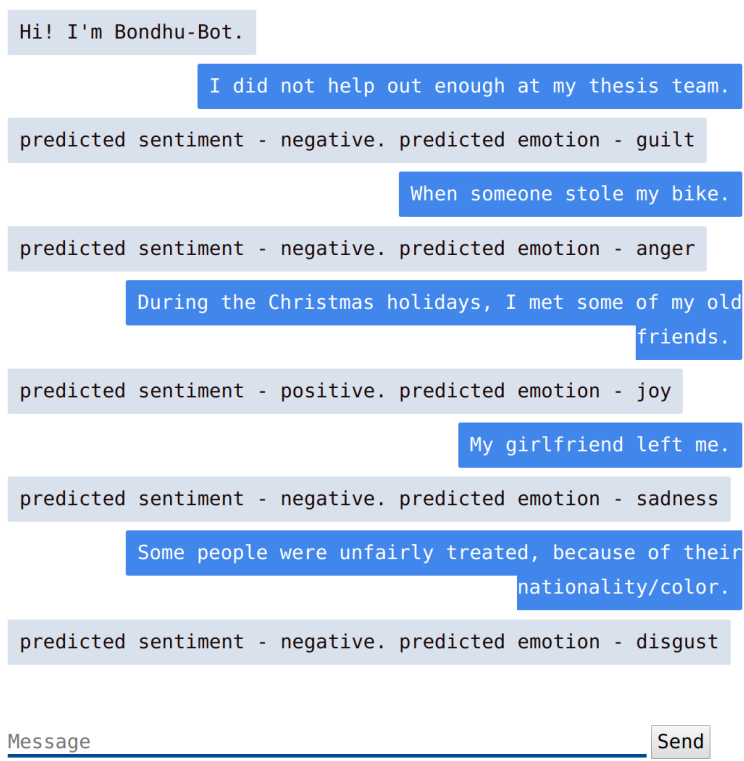

Figure 5.  Application of Emotion Analysis with Sentiment

## 4.3.  Intent Classification

Despite we got very accurate results through emotional analysis using Bi-directional GRU and Bi-directional LSTM, but for making the chatbot responses more human-like, knowing the intent of the message is essential. By emotion detection we will get the state of mind of the user, but *"why is he in that particular state?"* will be identified by knowing the intent behind it. Besides, knowing the context of a person's text helps to gather various information. As conversational agents are being used for various purposes like customer support, e-commerce and even in the field of mental health, determining the intention of a user has become very essential.

### 4.3.1. Data Processing

We used the dataset banking-77 which is composed of online banking queries and 78 corresponding intents [17]. After cleaning and refining the dataset, stemming was applied in the similar process like emotional analysis. Lemmatization groups all the similar words such as 'happy', 'happier', 'happiest' to get the ultimate source word even if it is written differently. Like before the One-Hot encoder was used to convert the categorical into numerical form. Similar tools used for one-hot encoder in emotion detection were also used in intent classification. Figure 6 describes the steps we followed for data processing.

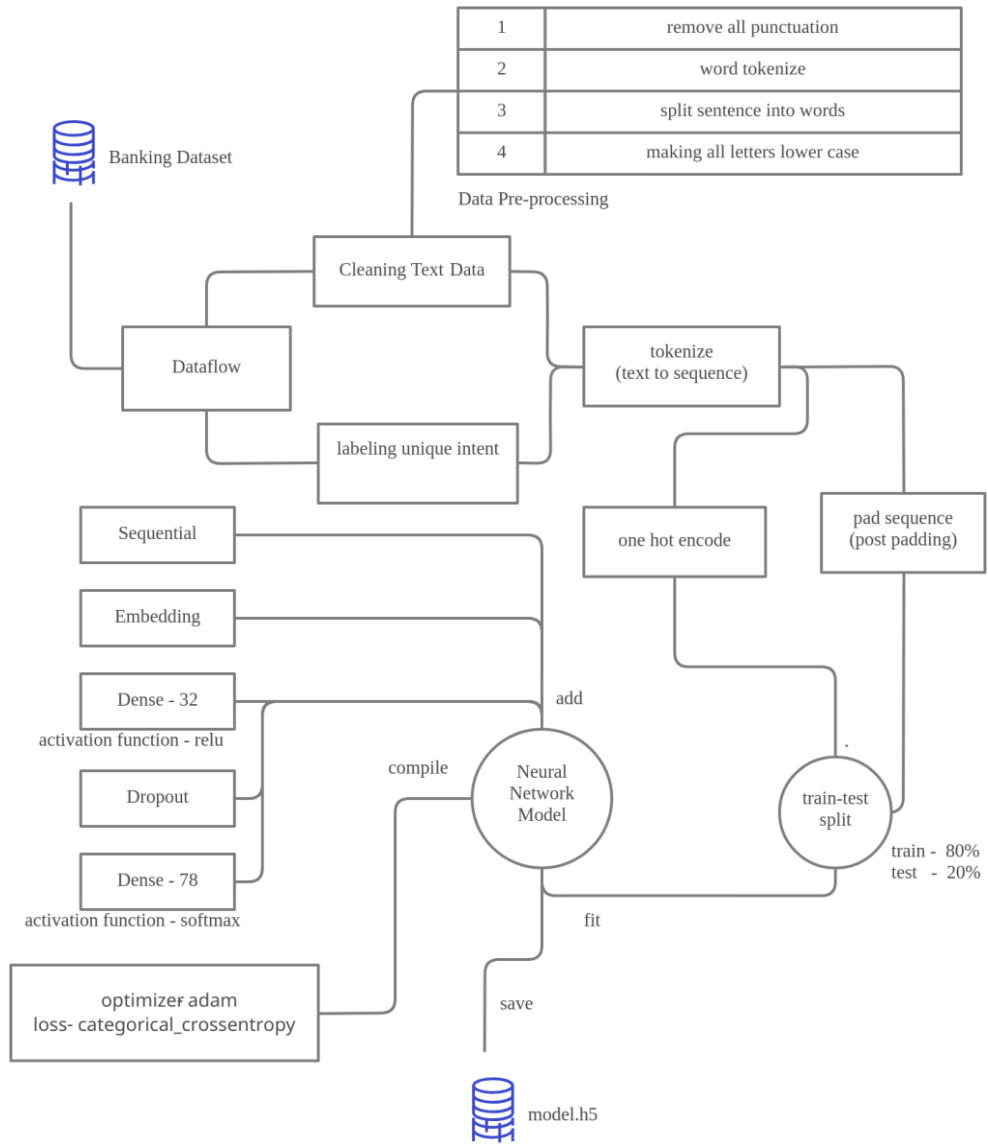

Figure 6.  Training Work Flow for Intent Classification

### 4.3.2. Method

We split our dataset into 80% of training dataset and 20% validation. Now, our data is ready to be fed in the model. We used Bi-GRU and Bi-LSTM, as it creates a reverse copy of the sentencesand reads from both the sides to keep both past and future context of sentence into consideration. All the parameters like activation functions, optimizer, batch size and epoch were set just like our previous analysis.

### 4.3.3. Result

From Table 6 we got the training accuracy of around 80.7% using Bi-GRU and 77.4% accuracy using Bi-LSTM which indicates a good comparison between the two models. However, Bi-GRU gives comparatively better results for intent classification.

Table 6. Accuracy for training and validation data of different Deep-Learning Models for Intent Classification

| Model | Training | | Validation | |
|---|---|---|---|---|
| | loss | accuracy | loss | accuracy |
| Bi-directional LSTM | 0.651 | 0.774 | 1.337 | 0.711 |
| Bi-directional GRU | 0.562 | 0.807 | 1.468 | 0.694 |

### 4.3.4. Performance analysis

Just as before, we gave some sentences as input for real time prediction. We created a get_intent class that gives the possible probabilities of similarities of the sentence with the intent categories. The intent that gives the max probability is the resultant outcome. As our intent dataset is related to bank queries, our input sentences in Figure 7 are different statements and questions regarding bank issues. In our second example we can see the user's card isn't working and our bot predicted *"card_not_working"* correctly. Similarly, the intents of the following examples were identified without any mistake.

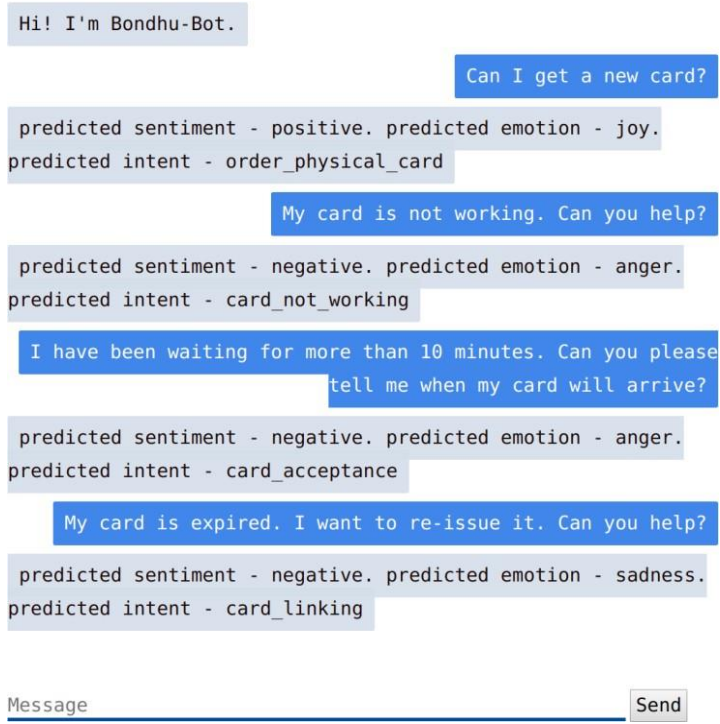

Figure 7.  Application of Intent Analysis with Sentiment and Emotion

## 4.4. Named-Entity Recognition

Now, our next work is to get details of the intent. By *'details'* we mean about whom or where or at which date and time the context of the sentence is referring to. It is important because entities such as person, location, organization, time etc. help to gather information related to the intent of the text. This kind of task is achieved by named-entity recognition. NER is a subtask in information extraction and machine translation and also various DRNN (Deep Recurrent Neural Network) models along with word embedding are applied to perform NER [18]. With the help of NER, we can identify and categorize key entities in text which will help our chatbots to become more interactive.

### 4.4.1. Data Processing

The dataset we used is Annotated Corpus for NER using GMB (Groningen Meaning Bank) corpus which is tagged and built for predicting entities such as name location, time etc[19]. Next, for processing the data, we have retrieved sentences and corresponding tags. As displayed in Figure 8 mappings between sentences and tags were defined by converting the words and tags to indexes.

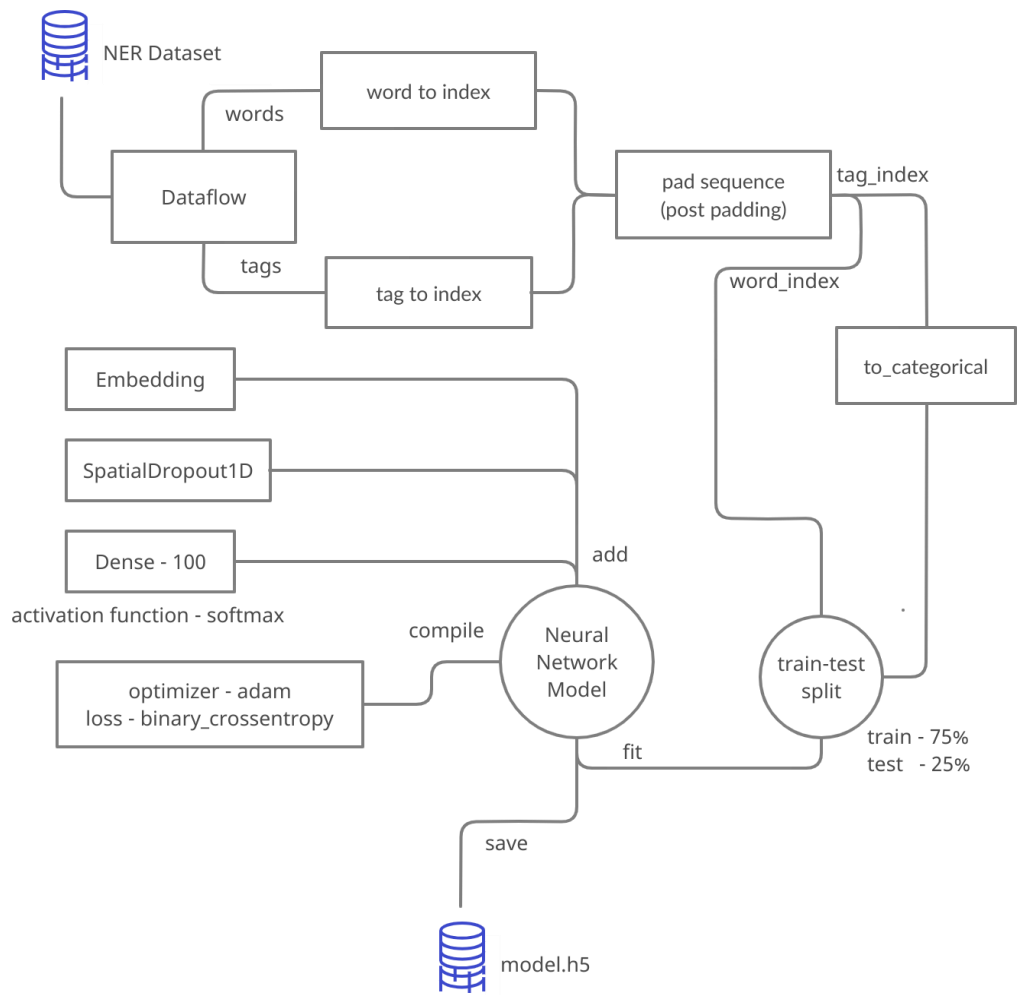

Figure 8. Training Work Flow for Named-entity Recognition

#### 4.4.2. Method

Now, we have created train and test splits so that we can estimate the performance of algorithms when they are used to make predictions on data. The training data was split to 90% and test data was split to 10%. Our next task was to build and compile bi-directional LSTM. We have applied Spatial dropout 1D to 0.1, to drop the entire 1D feature map. We also have kept the dense to 100 in our model. For training the model properly, we have used the ModelCheckpoint and livelossplot. The batch size is set to 32 and epoch set to 3. Finally, we have done the training and evaluation of the model.

#### 4.4.3. Result

After training the model, we got a training accuracy of 98.9% using both Bi-LSTM and Bi-GRU, given in Table 8, which plays a great role in helping chatbot to identify the entities and categorize them.

Table 8. Accuracy for training and validation data of different Deep-Learning Models forNamed-Entity Recognition

| Model | Training | | Validation | |
|---|---|---|---|---|
| | loss | accuracy | loss | accuracy |
| Bi-directional LSTM | 0.039 | 0.988 | 0.048 | 0.985 |
| Bi-directional GRU | 0.034 | 0.989 | 0.045 | 0.986 |

#### 4.4.4. Performance analysis

For real time predictions, we gave some sentences as input an example is shown in Table 9. In our input sentence *'George'* is the name of the person, *'London'* and *'Indonesia'* are the places he will be travelling within and *'sunday morning'* is the date and time of travel. All these entities were identified by our chatbot accurately.

Table 9. Predicted Named-entity for a sentence

| **Text:** George will go to London from Indonesia Sunday morning. | |
|---|---|
| **Entity** | **Predicted Tag** |
| George | I-per |
| London | B-geo |
| Indonesia | B-geo |
| Sunday | B-tim |
| Morning | I-tim |

Finally, in Figure 9, we incorporated all the features analysed above. We used variations in input text which in return gave us different sentiments, emotion, intent and entity. Considering the intent dataset, we needed to set examples confined to banking. However, our first input was nowhere related to bank issues but surprisingly it gives *"edit_personal_details"* which was not as bad as we had expected for this particular text. In addition, all other components for the given text were accurately identified such as *'sad'* emotion, *'negative'* sentiment and B-time *'Sunday'*. The second input was *"I lost my phone yesterday, so I could not help out enough to my thesis team"* which is a complex sentence. But our chatbot could predict all the components of the complex sentence and predicted sentiment as *'negative'*, emotion as *'guilt'*, intent as *"lost_or_stolen phone"* and entity yesterday as *'time'*. Next, we gave a bit more complex sentence which had multiple entities and it predicted all four entities. The examples also show us that our bot is now completely prepared to understand a user's given dialogue. This can be used in different sectors for understanding queries or statements of a chatbot user simply by changing the intent dataset to the desired intent domain.

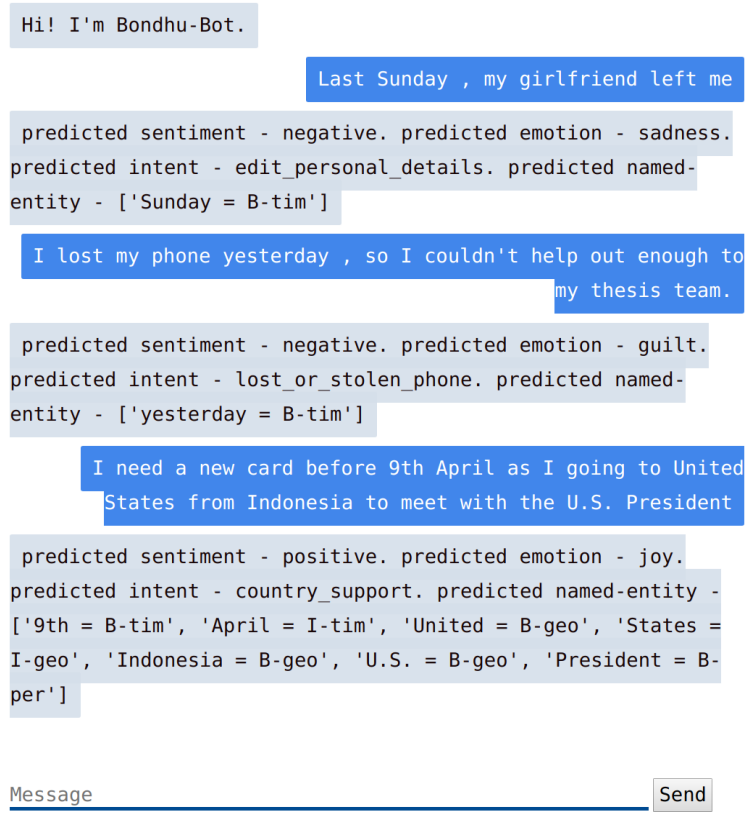

Figure 9. Application of Name-Entity Analysis with Sentiment, Emotion and Intent

## 5. COMPARISON

Initially, we used a machine learning algorithm - **Support Vector Machine**, to evaluate performance on sentiment analysis and emotion detection. The reason to use SVM was its mathematical foundation in statistical learning theory that made it popular in the field of machine learning and supervised classification [20]. Moreover, it was also used to recognize, interpret and process human emotion from text according to predefined emotion classes [21] Figure 10 shows results from sentiment analysis and emotional detection with SVM were not very satisfactory. It gave only 48.5% accuracy for sentiment analysis and only 14.16% for emotion detection where else all our experimental accuracy using deep learning varied between 74% to 98%. Thus, from the results we can deduce our deep learning approach outperformed ML algorithm by a significant percentage.

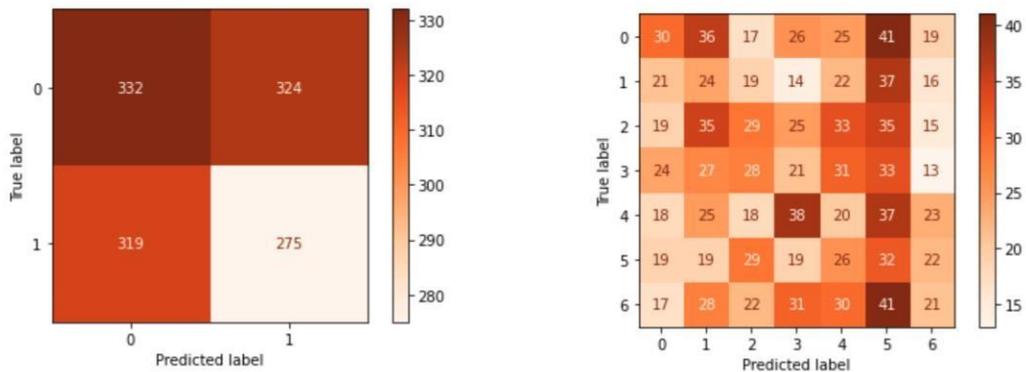

(a) confusion matrix for sentiment analysis          (b) confusion matrix for emotion detection

Figure 10.  Performance evaluation using Support Vector Machine

## 6. CONCLUSION

Throughout our research, we extracted different components from texts that will help a system to have adequate information about the user's opinion or statement. This paper proposes that for a chatbot to appropriately respond to users it needs to identify the important keywords to capture the expressed sentiments, emotions as well as the intention, behind the uttered dialogue of the user in depth. We analysed all the major components through application of deep learning techniques and achieved our goals in making accurate predictions in each segment of work. However, through our work we understood that it is very difficult to build a chatbot that can mimic a human and make conversations interactive. The detailed analysis incorporating all the functions demonstrates that our proposed model will significantly help to build a human-like chatbot with in numerous applications.

## 7. FUTURE WORK

As now our bot is already able to acknowledge the personalised words and emotions, in future anyone including us can be able to build a complete responsive chatbot using this which can communicate in a more humanistic manner. If the chatbot is trained with a dialogue corpus dataset of any specific domain, it will be able to respond referring to all the analysed features we worked on in this research paper. All it needs is to add a dialogue generation feature in our model so thatit can generate a response automatically to the given input text by the user. To solve the unavailability of appropriate dataset and to make the chatbot more human-like, we have the plan to extend our approach to unsupervised learning in the future. And then, our bot will be able to train itself on previous history of conversations and infer the dependencies between input texts.

## REFERENCES

[1]   Solutions, R. I. (2018, April 13). Top 10 Reasons: Why Your Business Need A Chatbot Development. Retrieved January 4, 2021, from https://chatbotsjournal.com/top-10-reasons-why-your-business- need-a-chatbot-development-5a53760da1b6?gi=491b46b2ea69

[2]   Advances    in    Conversational    AI.    (n.d.).    Retrieved    January    4,    2021,    from https://ai.facebook.com/blog/advances-in-conversational-ai/

[3]   Deep Learning vs Machine Learning: Know the Difference. Retrieved January 4, 2021, from https://www.mygreatlearning.com/blog/is-deep-learning-better-than-machine-learning

[4]    Kawade, D. R., & Oza, K. S. (2017). Sentiment analysis: machine learning approach. Int J Eng Technol, 9(3), 2183-2186.

[5]    Hossain, Y., Hossain, I. A., Banik, M., & Chakrabarty, A. (2018, June). Embedded system based Bangla intelligent social virtual robot with sentiment analysis. In 2018 Joint 7th International Conference on Informatics, Electronics & Vision (ICIEV) and 2018 2nd International Conference on Imaging, Vision & Pattern Recognition (icIVPR) (pp. 322-327). IEEE.

[6]    Lee, D., Oh, K. J., & Choi, H. J. (2017, February). The chatbot feels you-a counseling service using emotional response generation. In 2017 IEEE international conference on big data and smart computing (BigComp) (pp. 437-440). IEEE.

[7]    Bhagwat, V. A. (2018). Deep Learning for Chatbots.

[8]    Acheampong, F. A., Wenyu, C., & Nunoo-Mensah, H. (2020). Text-based emotion detection: Advances, challenges, and opportunities. Engineering Reports, 2(7), e12189.

[9]    Balaji, M., & Yuvaraj, N. (2019, November). Intelligent Chatbot Model to Enhance the Emotion Detection in social media using Bi-directional Recurrent Neural Network. In Journal of Physics: Conference Series (Vol. 1362, No. 1, p. 012039). IOP Publishing.

[10]   Yadav, V., & Bethard, S. (2019). A survey on recent advances in named entity recognition from deep learning models. arXiv preprint arXiv:1910.11470.

[11]   Meng, L., & Huang, M. (2017, November). Dialogue intent classification with long short-term memory networks. In National CCF Conference on Natural Language Processing and Chinese Computing (pp. 42-50). Springer, Cham.

[12]   Kharde, V., & Sonawane, P. (2016). Sentiment analysis of twitter data: a survey of techniques. arXiv preprint arXiv:1601.06971.

[13]   Maas, A., Daly, R. E., Pham, P. T., Huang, D., Ng, A. Y., & Potts, C. (2011, June). Learning word vectors for sentiment analysis. In Proceedings of the 49th annual meeting of the association for computational linguistics: Human language technologies (pp. 142-150).

[14]   David. (2017, November 30). Twitter_sentiment [Data Set]. Retrieved from https://www.kaggle.com/ywang311/twitter-sentiment

[15]   Pennington, J., Socher, R., & Manning, C. D. (2014, October). Glove: Global vectors for word representation. In Proceedings of the 2014 conference on empirical methods in natural language processing (EMNLP) (pp. 1532-1543).

[16]   Scherer, K. R., & Wallbott, H. G. (1994). Evidence for universality and cultural variation of differential emotion response patterning. Journal of personality and social psychology, 66(2),310.

[17]   Casanueva, I., Temčinas, T., Gerz, D., Henderson, M., & Vulić, I. (2020). Efficient intent detection with dual sentence encoders. arXiv preprint arXiv:2003.04807.

[18]   Khan, W., Daud, A., Alotaibi, F., Aljohani, N., & Arafat, S. (2020). Deep recurrent neural networks with word embeddings for Urdu named entity recognition. ETRI Journal, 42(1), 90-100.

[19]   Walia, A. (2017, September 21). Annotated Corpus for Named Entity Recognition [Data Set]. Retrieved from https://www.kaggle.com/abhinavwalia95/entity-annotated-corpus

[20]   Awad, M., & Khanna, R. (2015). Support vector machines for classification. In Efficient Learning Machines (pp. 39-66). Apress, Berkeley, CA.

[21]   Kirange, D. K., & Deshmukh, R. R. (2012). Emotion classification of news headlines using SVM. Asian Journal of Computer Science and Information Technology, 5(2), 104-106.